\documentclass{article}

\PassOptionsToPackage{numbers, compress}{natbib}
\usepackage[preprint]{neurips_2026}
\usepackage[utf8]{inputenc}
\usepackage[T1]{fontenc}
\usepackage{hyperref}
\usepackage{url}
\usepackage{booktabs}
\usepackage{amsfonts}
\usepackage{amsmath}
\usepackage{amssymb}
\usepackage{microtype}
\usepackage{graphicx}
\usepackage{comment}
\usepackage{multirow}
\usepackage[table,xcdraw]{xcolor}

\usepackage{ulem}
\usepackage{algorithm}
\usepackage{algorithmicx}
\usepackage{algpseudocode}
\usepackage{float}
\usepackage{mdframed}
\usepackage{fvextra}
\usepackage{wrapfig}
\usepackage{subcaption}
\usepackage[most]{tcolorbox}
\newtcolorbox{appendixpromptbox}{
  enhanced,
  breakable,
  colback=yellow!10,
  colframe=black,
  boxrule=0.5pt,
  arc=4pt,
  left=8pt,
  right=8pt,
  top=8pt,
  bottom=8pt
}
\title{DiagramRAG: A Lightweight Framework to Retrieve Scientific Diagram for Figure
Generation}

\author{
\textbf{Xinjiang Yu}\thanks{Equal contribution.}
\quad
\textbf{Junyi Han}\footnotemark[1]
\quad
\textbf{Zhuofan Chen}\footnotemark[1]
\quad
\textbf{Chi Zhang}
\quad
\textbf{Xiangyu Fu}
\\
\textbf{Jingyuan Tan}
\quad
\textbf{Zirui You}
\quad
\textbf{Yixiang Jian}
\quad
\textbf{Yu-Ping Wang}
\quad
\textbf{Chengliang Chai}\thanks{Corresponding author: ccl@bit.edu.cn}
\\
Beijing Institute of Technology
\\
\texttt{\{3220251215, hanjunyi\}@bit.edu.cn}
}
\begin{document}

\maketitle

\begin{abstract}
Scientific diagrams are essential for communicating complex methodologies in academic papers. A natural way for researchers to specify such diagrams is through rough sketches, where text labels, connectors, and spatial arrangements express early semantic and topological intentions. However, sketches are usually incomplete, making them insufficient for directly producing publication-quality diagrams. Existing sketch-based generation methods mainly reconstruct the sketch itself, while recent text-driven diagram generation frameworks rely on textual semantics and do not fully exploit the topological structure contained in sketches. In this paper, we introduce \texttt{DiagramRAG}, a lightweight retrieval-augmented framework for sketch-based scientific diagram completion. Given a user sketch, \texttt{DiagramRAG} retrieves reference diagrams that are both semantically relevant to the sketch content and topologically compatible with its structure, and uses them to guide downstream diagram generation. To enable efficient structure-aware retrieval, we represent diagrams as knowledge graphs, synthesize sketch variants at different simplification levels, and train an embedding model to align sketches with compatible diagrams in a shared space. The retrieved references further provide content, topology, and visual priors for completing and rendering the final diagram. Experiments show that \texttt{DiagramRAG} achieves F1-scores of 0.848 and 0.802 on \texttt{DiagramBank} and \texttt{FigureBench}, respectively, and improves generation quality with the best VLM-as-a-Judge score of 7.170, while reducing inference latency to 35.48 seconds per sample. Our code and data are available at \url{https://anonymous.4open.science/r/DiagramRAG-A262} and \url{https://huggingface.co/datasets/anonymous-review-a262/DiagramSketch}.
\end{abstract}

\section{Introduction}

Scientific diagrams play a central role in academic communication, often serving as structural-rich visual summaries~\cite{kembhavi2016diagram,kim2018dynamic,zhang2026diagrambank} of complex methodologies~\cite{drucker2014graphesis,tversky2011visualizing,lee2018viziometrics}. 
Despite recent advances in automated diagram generation~\cite{paperbanana2026,autofigure2026}, producing high-quality diagrams remains a labor-intensive process~\cite{huang2026scifig}. 
In practice, researchers often rely on sketches to externalize their ideas, since sketches can naturally express semantic content through text labels and topological structure through connectors and spatial arrangement~\cite{tversky2011visualizing,saito2025sketch2diagram,ellis2018learning}. 
Nevertheless, such semantic and topological cues are usually sparse and abstract, making sketches insufficient for producing publication-quality scientific diagrams~\cite{tan2025sketchagent,saito2025sketch2diagram}. 
Therefore, sketches require further completion and rendering, which makes sketch-based diagram completion an important task: a system should not simply refine the visual appearance of the input sketch, but should also complete its information with semantically and topologically compatible references.

Existing work has made progress in generating diagrams from sketches, but completing rough sketches into publication-quality diagrams still remains challenging. Sketch-based generation methods~\cite{saito2025sketch2diagram,zhang2023controlnet} such as \texttt{SketchAgent}~\cite{tan2025sketchagent} convert hand-drawn sketches into structured representations by parsing diagram elements and relations. 
However, these methods primarily focus on reconstructing the input sketch itself, rather than enriching its content and topology.
Another possible method~\cite{paperbanana2026,autofigure2026,huang2026scifig,shrivastava2024diagrammergpt} is to convert sketches into textual descriptions and then employ multi-agent text-to-diagram generation frameworks. 
Among these frameworks, \texttt{PaperBanana}~\cite{paperbanana2026} uses VLMs to retrieve reference examples~\cite{lewis2020rag,zhang2026diagrambank} based on textual information and coordinates specialized agents to generate diagrams. 
However, such retrieval primarily relies on textual semantics and does not fully exploit the topological structure captured in sketches~\cite{johnson2015image,krishna2017visualgenome,lu2016visual,xu2017scenegraph,zellers2018neural}. Moreover, VLM-based reference retrieval~\cite{radford2021clip,openclip,li2026qwen3} incurs substantial computational cost, making it difficult to scale to large diagram collections.

Motivated by these limitations, we propose \texttt{DiagramRAG}, a lightweight framework that completes sketches by retrieving reference diagrams with semantic relevance and topological compatibility. Given a user sketch, our goal is to retrieve semantically relevant and topologically compatible diagrams from a curated diagram set as references to guide the generation of publication-quality diagrams. To achieve this, \texttt{DiagramRAG} first curates a high-quality diagram set from raw diagram collections~\cite{autofigure2026,zhang2026diagrambank} and represents diagrams as knowledge graphs~\cite{kembhavi2016diagram,kim2018dynamic,johnson2015image} that capture both semantic content and topological structure. Using these representations, we generate sketch variants at different simplification levels to construct sketch-diagram supervision pairs. We then train a lightweight embedding model~\cite{sain2023clip,pr2021,bai2019simgnn} using contrastive objectives to learn a structure-aware retrieval space, where sketches are aligned with compatible reference diagrams. 
For efficient retrieval, we precompute diagram embeddings into an offline index. At query time, the user sketch is encoded into the same space, and top-$k$ references are retrieved via similarity search. The retrieved references provide content, structural, and visual guidance for downstream generation, enabling \texttt{DiagramRAG} to produce diagrams that are structurally complete, semantically enriched, and visually polished.

We summarize our main contributions as follows:
\begin{itemize}
\item We formulate sketch-based diagram completion, where rough user sketches are transformed into publication-quality diagrams with the support of retrieved reference diagrams.
\item We propose a lightweight structure-aware retrieval framework that constructs sketch-diagram supervision and trains a contrastive embedding model to align sketches with structurally compatible diagrams while preserving semantic relevance.
\item We validate \texttt{DiagramRAG} through comprehensive retrieval and generation experiments on \texttt{DiagramBank} and \texttt{FigureBench}, showing that it enhances VLM-as-a-Judge generation quality, improves structure-aware retrieval, and achieves efficient inference for publication-ready diagram generation.
\end{itemize}

\section{Related Work}
\textbf{Automated Scientific Diagram Generation.} Recent progress in automated scientific diagram generation can be broadly grouped into three lines of work. The first line is text-driven scientific figure generation, where systems generate diagrams from textual inputs. Representative methods, such as \texttt{PaperBanana}~\cite{paperbanana2026} and \texttt{AutoFigure}~\cite{autofigure2026} generate intermediate layouts, retrieve textual or visual references, and refine outputs through planning or multi-agent pipelines. While these methods~\cite{shrivastava2024diagrammergpt,huang2026scifig} have significantly improved automation, they largely rely on textual descriptions. The second line starts from sketches. Methods such as \texttt{Sketch2Diagram}~\cite{saito2025sketch2diagram} and \texttt{SketchAgent}~\cite{tan2025sketchagent} focus on converting hand-drawn sketches into vector code or structured, machine-readable diagrams. However, they mainly emphasize structural parsing and reconstruction, rather than information completion. The third line includes controllable image generation methods such as \texttt{ControlNet}~\cite{zhang2023controlnet}, which inject low-level spatial conditions into diffusion models. Although these methods provide stronger control over layout cues, they are mainly designed to preserve local spatial patterns.

\textbf{Multimodal Retrieval-Augmented Generation (RAG).}  Retrieval-augmented generation (RAG) incorporates external knowledge into generation and has been extended to multimodal settings~\cite{lewis2020rag,radford2021clip}. Prior work~\cite{sarthi2024raptor,edge2025graphrag} explores structure-driven retrieval strategies for text in RAG systems. Besides, diagram-oriented resources for RAG reflect a growing trend of treating scientific diagrams as retrievable knowledge sources~\cite{zhang2026diagrambank} .However, existing MRAG methods~\cite{radford2021clip,khattab2020colbert} primarily focus on semantic relevance and do not explicitly model structural compatibility, which is critical for sketch-based diagram retrieval with limited textual information.

\textbf{Diagram Similarity and Structure Matching.} Recent work increasingly models diagrams as structured objects. Benchmarks such as \texttt{DiagramEval}~\cite{liang2025diagrameval} and \texttt{SciFlow-Bench}~\cite{zhang2026sciflowbench} evaluate diagrams via graph representations. Prior approaches include shared embedding models for sketch-image matching~\cite{radford2021clip,sain2023clip} and graph-based similarity methods~\cite{bai2019simgnn,becigneul2020otgnn,neuhaus2005bridging} such as \texttt{GNN} and \texttt{GED}, along with subgraph matching techniques~\cite{sun2024comprehensive,han2025comprehensive}. However, these methods typically assume explicit graph inputs or focus on symmetric similarity, rather than retrieval from raster sketches with completion-oriented structural compatibility.

\section{Method}

In this section, we first define the task and introduce the core notations in Sec.~\ref{sec:problem definition}. 
We then present the overall framework in Sec.~\ref{sec:overall framework}, which consists of three components: Structure-aware Representation Learning in Sec.~\ref{sec:structure-aware representation learning}, Structure-aware Retrieval in Sec.~\ref{sec:structure-aware retrieval}, and Refined Diagram Generation in Sec.~\ref{sec:refined diagram generation}.

\begin{figure}[htbp]
  \centering
  \includegraphics[width=\linewidth]{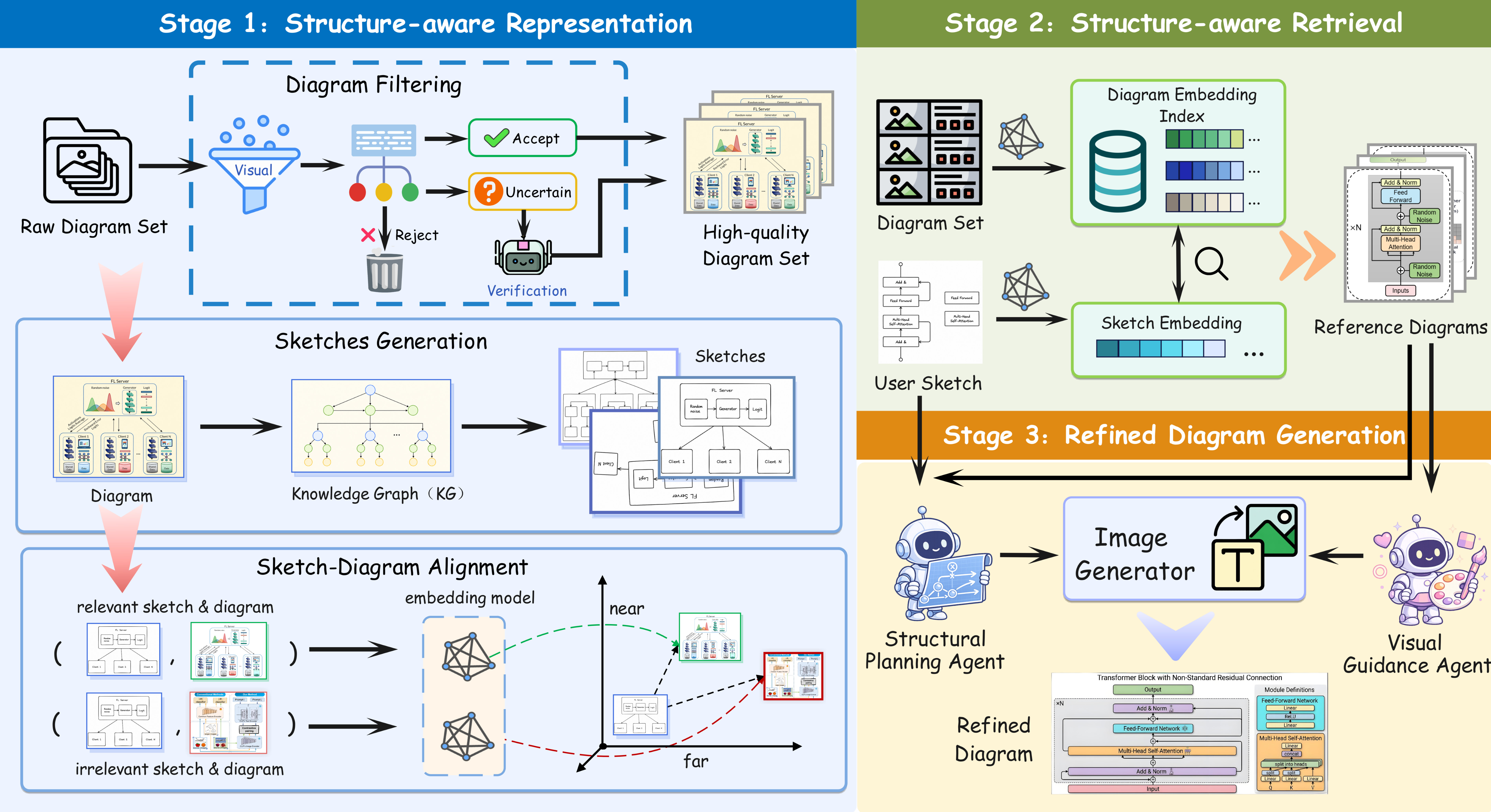}
  \caption{Overview of 
  \texttt{DiagramRAG}.}
  \label{fig:framework}
\end{figure}

\subsection{Problem Definition}
\label{sec:problem definition}
In this paper, we study the problem of sketch-based diagram completion over a high-quality diagram set $\mathcal{D}$, where each diagram $d \in \mathcal{D}$ is a publication-quality diagram and the sketch $s$ is a simplified diagram representation. The objective is to retrieve diagrams that are semantically relevant and topologically compatible with $s$ and generate a refined diagram $I$ based on them. In the offline training stage, we learn encoders that map sketches and diagrams into a shared embedding space via contrastive learning, where the similarity $\mathrm{sim}(s,d)$ is measured based on their semantic and topological compatibility; we compute the sketch embedding and directly compare it against the precomputed diagram embedding index, retrieving the top-$k$ most similar diagrams to form the reference set $R$, based on which a generator produces the final refined diagram ${I}$ by enriching the semantic content and topological structure of the sketch and rendering it into a visually refined diagram $I$.

\subsection{Overall Framework}
\label{sec:overall framework}

As illustrated in Figure~\ref{fig:framework}, \texttt{DiagramRAG} follows a three-stage pipeline for sketch-based diagram completion. In the first stage, \texttt{DiagramRAG} constructs structure-aware training data from a raw diagram set by first selecting a high-quality subset of diagrams. Starting from these curated diagrams, we extract knowledge graph (KG) representations and generate diverse sketch variants to simulate different levels of simplification, enabling the model to learn representations that capture both semantic content and topological structure. Using the generated sketches together with their original diagrams, we construct positive and negative sketch-diagram pairs for contrastive learning. This trains the model to differentiate diagrams that are semantically and topologically consistent with a given sketch from those that are not, thereby aligning the sketch space with the diagram space. In the second stage, given a user sketch, \texttt{DiagramRAG} encodes it into the learned embedding space and retrieves the top-$k$ compatible diagrams from a precomputed diagram embedding index as references. In the final stage, \texttt{DiagramRAG} leverages the retrieved diagrams to guide generation. A structural planning agent infers a completion plan from the sketch and references, while a visual guidance agent extracts aesthetic patterns from the references. These signals are combined to condition the downstream generator, producing the final publication-quality diagram that is structurally complete, semantically enriched, and visually refined.

The remainder of this section is organized as follows. We first introduce structure-aware representation learning in Sec.~\ref{sec:structure-aware representation learning}. We then describe the structure-aware retrieval mechanism in Sec.~\ref{sec:structure-aware retrieval}. Finally, we present the reference-guided diagram generation process in Sec.~\ref{sec:refined diagram generation}.

\subsection{Structure-aware Representation Learning}
\label{sec:structure-aware representation learning}
\textbf{Motivation.} Sketches provide compact visual specifications, where textual labels convey semantic content and spatial layout reveals topological structure. However, existing sketch interpretation methods based mainly on textual semantics makes
limited use of the rich topological information inherently present in sketches, including connectivity, containment, and spatial layout. This motivates a structure-aware representation that preserves semantic cues while explicitly organizing diagram elements and their relations. To this end, we represent diagrams as knowledge graphs and construct supervised sketch-diagram training data by synthesizing sketch variants with different degrees of simplification. These supervision pairs provide the foundation for learning sketch-diagram alignment in the subsequent retrieval stage. 
As illustrated in Figure~\ref{fig:framework}, this stage consists of diagram filtering, sketch generation, and sketch-diagram alignment.

\textbf{Diagram Filtering.} The raw diagram set extracted from scientific publications~\cite{autofigure2026,zhang2026diagrambank} is highly heterogeneous, containing a large proportion of plots and natural images, as well as samples with insufficient structural information and poor visual quality, making it unsuitable for structure-aware representation. So, we construct a high-quality diagram set $\mathcal{D}$ through a multi-stage filtering pipeline. First, we extract low-level 
visual features from the diagram set using conventional image processing techniques, and perform coarse filtering to remove samples with insufficient 
structural content or poor visual quality. Next, we jointly model high-level semantic alignment and visual quality by leveraging 
CLIP-based image representations~\cite{radford2021clip} together with low-level visual features. Specifically, 
we employ a pre-trained LightGBM classifier~\cite{ke2017lightgbm} to estimate a retention probability for 
each image based on these combined features. According to predefined thresholds, 
candidate images are categorized into accepted, rejected, and uncertain groups. Furthermore, for samples in the uncertain group, we employ a vision-language model~\cite{lee2024prometheusvision} to perform fine-grained verification under explicit task criteria, focusing on whether 
the image belongs to structured diagram types suitable for \texttt{DiagramRAG}. Through this pipeline, we obtain a high-quality diagram set $\mathcal{D}$ with clear 
structure, consistent content, and strong suitability for downstream tasks. Implementation details and feature definitions are provided in Appendix~\ref{app:diagram_filtering}.

\textbf{Sketch Generation.}  To enable controllable sketch synthesis, we first represent each diagram as a diagram-specific knowledge graph (KG), rather than directly operating on pixels. 
This design is inspired by prior works~\cite{johnson2015image,krishna2017visualgenome,kembhavi2016diagram,kim2018dynamic}on scene graphs and diagram parse graphs (DPGs), which shows that visual content can be abstracted into explicit objects, attributes, and relations for retrieval and reasoning. 
In our setting, such a structured representation is particularly suitable for scientific diagrams, since textual labels, connectors, containment relations, and spatial layouts jointly define the semantic content and topological structure of a diagram. 
By making these components explicit, the KG enables us to synthesize sketch variants through controlled graph-level simplification and perturbation, while preserving the core topological relations of the original diagram.

To efficiently construct training supervision, we sample a subset 
$\mathcal{D}_s \subset \mathcal{D}$ instead of processing the entire diagram set. 
This reduces the computational overhead of structural parsing while preserving sufficient structural diversity for learning. 
For each diagram $d \in \mathcal{D}_s$, we first parse it into a KG representation and then generate a set of sketches $\mathcal{S}_d$ by applying controlled graph-level transformations in the KG space. 
In this representation, nodes describe diagram elements through attributes such as element type and textual name, while edges capture relational structure through source-target connections and relation labels. Each sketch $s \in \mathcal{S}_d$ serves as a simplified variant of its source diagram $d$, reflecting common properties of user-drawn sketches: incomplete structural components, sparse textual labels, and imprecise spatial layouts. To capture these properties, we apply three types of graph-level transformations: importance-aware merging of locally and semantically related nodes, textual-label suppression, and spatial-layout perturbation, producing sketches with varying degrees of structural simplification, textual sparsity, and drawing noise while retaining the essential topology of the source diagram.

\begin{figure}[htbp]
  \centering
  \includegraphics[width=\linewidth]{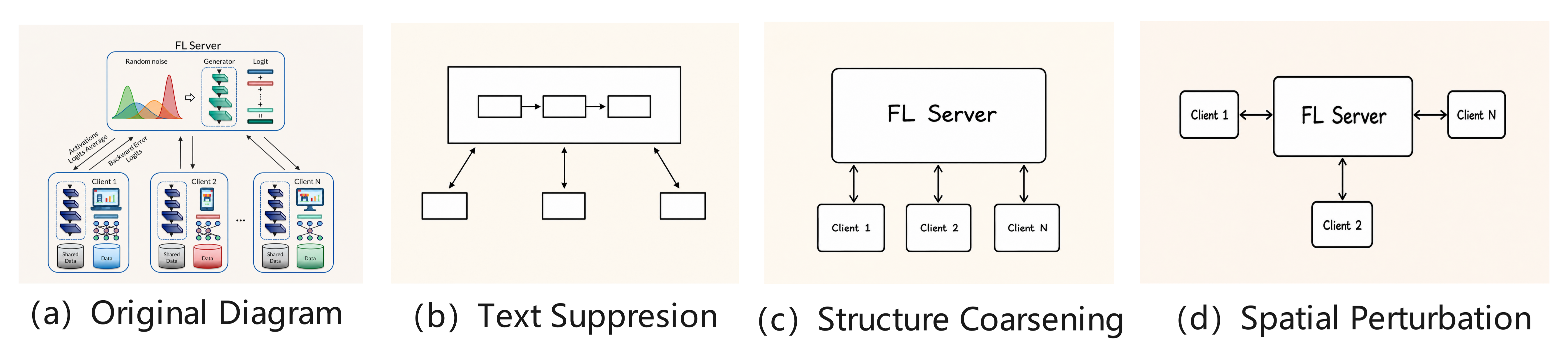}
  \caption{Examples of generated sketch variants}
  \label{fig:sketch variants}
\end{figure}

Figure~\ref{fig:sketch variants} illustrates specific examples of sketches generated from the KG representation. Starting from the original diagram in Figure~\ref{fig:sketch variants} (a), we present three representative variants: the text-suppressed sketch in Figure~\ref{fig:sketch variants} (b) removes most textual labels while preserving the original layout; the structure-coarsened sketch in Figure~\ref{fig:sketch variants} (c) compresses fine-grained nodes into a compact structural backbone; and the spatially perturbed sketch in Figure~\ref{fig:sketch variants} (d) rearranges the coarse components to reflect imperfect hand-drawn layouts. These examples show how multiple sketch variants can be derived from a single diagram, varying in textual completeness, structural granularity, and layout precision, thereby providing diverse supervision for structure-aware matching.

\textbf{Sketch-Diagram Space Alignment.} We learn a structure-aware embedding space for sketches and diagrams via contrastive learning. In our 
setting, we construct supervision by pairing each synthesized sketch with its source diagram, 
and train the embedding model to capture their structural correspondence.

Let $f_s(\cdot)$ and $f_d(\cdot)$ denote the sketch encoder and diagram encoder with the same set of weight. 
For each synthesized sketch $s$, we treat its source  diagram as the 
positive sample $d^{+}$ . Negative samples $d^{-}$  are obtained through in-batch sampling, where diagrams from 
other sketch-diagram pairs in the same mini-batch are used as negatives. The similarity between a sketch $s$ and a diagram $d$ is measured by cosine similarity:
\begin{equation}
\mathrm{sim}(s,d)=
\frac{
f_s(s)\cdot f_d(d)
}{
\|f_s(s)\| \, \|f_d(d)\|
}.
\end{equation}

The model is trained with the following contrastive objective:

\begin{equation}
\mathcal{L} =
-\log
\frac{
\exp(\mathrm{sim}(s, d^{+})/\tau)
}{
\exp(\mathrm{sim}(s, d^{+})/\tau)
+
\sum_{d^{-} \in \mathcal{N}}
\exp(\mathrm{sim}(s, d^{-})/\tau)
},
\end{equation}

where $\mathcal{N}$ denotes the set of negative samples drawn from the current mini-batch, 
and $\tau$ is a temperature parameter. Under this formulation, sketches are encouraged to be close to their corresponding diagrams in the embedding space, while remaining separated from unrelated diagrams. 
As a result, diagrams that share relevant textual semantics and compatible structural patterns with a given sketch are mapped to nearby regions and can be retrieved as reference diagrams. 
This training strategy enables the model to capture KG-level correspondence between sketches and diagrams, allowing it to retrieve diagrams that are both semantically relevant and topologically compatible at inference time.

\subsection{Structure-aware Retrieval.}
\label{sec:structure-aware retrieval}
Given the structure-aware embedding model learned in Sec.~\ref{sec:structure-aware representation learning}, 
we use it to retrieve diagrams that are structurally compatible with a user-provided sketch.

Specifically, we first encode all diagrams in the high-quality set $\mathcal{D}$ using the 
diagram encoder $f_d(\cdot)$ and store their embeddings as a retrieval library 
$\{ f_d(d) \mid d \in \mathcal{D} \}$, which serves as a fixed index for efficient similarity-based retrieval. At inference time, given a sketch $s$, we compute its embedding using $f_s(\cdot)$ and measure 
its similarity with all diagrams using the similarity function defined in 
Sec.~\ref{sec:structure-aware representation learning}. We then retrieve the top-$k$ most similar diagrams:

\begin{equation}
R=\text{Top}k\big(\{\mathrm{sim}(s,d)\}_{d\in\mathcal{D}}\big).
\end{equation}

The retrieved set $R$ forms the reference diagram collection, consisting of diagrams that are semantically relevant to and topologically compatible with the input sketch. Notably, compared to approaches that perform explicit structure parsing or large vision-language models 
at inference time, our method is lightweight and efficient. Both diagram and sketch embeddings 
are computed using a locally deployed model, eliminating the need for repeated calls to expensive 
models for structure parsing. This design not only preserves structure-aware matching by 
aligning sketch and diagram embedding spaces, but also significantly reduces computational cost, 
enabling efficient and scalable deployment.

\subsection{Refined Diagram Generation}
\label{sec:refined diagram generation}

Building upon the semantically and structurally compatible reference diagrams retrieved in Sec.~\ref{sec:structure-aware retrieval}, this stage generates the final diagram by integrating explicit structure planning with visual guidance. Given an input sketch $s$ and the retrieved set $R=\{d_1,\dots,d_k\}$, we first employ a structural planning agent to produce a high-level structure prompt 
\begin{equation}
P = \mathrm{Plan}(s, R),
\end{equation}
that specifies the key components and their connectivity. This step leverages the structural patterns in $R$ to infer a topology that is both complete and compatible with the input sketch, effectively compensating for the sparsity and incompleteness of the sketch. In parallel, a Visual Guidance Agent summarizes aesthetic cues from $R$ into a style prompt 
\begin{equation}
G = \mathrm{Style}(R),
\end{equation}
providing guidance on layout, color, and composition. The structure prompt $P$ and style prompt $G$ are then jointly fed into a generative model (e.g., \texttt{Nano Banana Pro} or \texttt{GPT-Image 2}) to produce the final diagram 
\begin{equation}
I = \mathrm{Gen}(P, G).
\end{equation}
Under this framework, the generator performs semantic enrichment, topological completion, and visual rendering based on the retrieved references. Since the reference set $R$ is obtained through structure-aware retrieval, it provides both content and topology priors that guide the generation process. These priors help the generator enrich sparse sketch semantics, complete structural relations, and preserve coherent visual organization, thereby reducing semantic drift, structural hallucination, missing components, and incorrect connections.

Overall, the three stages together transform a simplified input sketch into a publication-quality diagram that is semantically aligned with the user's intent and topologically compatible with the input sketch. By retrieving compatible references and using them to guide generation, \texttt{DiagramRAG} enriches semantic content, completes structural relations, and produces a visually polished diagram suitable for scientific papers.

\section{Experiments}


\subsection{Experimental Setup}

\textbf{Datasets and Data Synthesis.} To evaluate \texttt{DiagramRAG}, we construct a diagram pool $\mathcal{D}$ based on the DiagramBank and FigureBench datasets. To ensure structural suitability and high visual fidelity, we apply the multi-stage filtering pipeline described in Sec.~\ref{sec:structure-aware representation learning} to the raw collection; full implementation details of this pipeline are provided in Appendix~\ref{app:diagram_filtering}. 
For training data construction, we first extract diagram-internal knowledge graphs using Qwen3.6-Plus. We then employ GPT-4o and Gemini 3.1 Flash to synthesize five distinct classes of sketch variants, designed to emulate varying degrees of structural incompleteness and layout perturbations. The full generation protocol, associated degradation statistics and the specific prompt templates for KG extraction and sketch synthesis for these variants are presented in Appendix~\ref{app:kg_variant_generation}, Appendix~\ref{app:kg_variant_losses}, and Appendix~\ref{app:prompt}, respectively. 
During contrastive training, we form triplets via a structured mini-batch sampling strategy: each micro-batch samples 20 target diagram IDs, and each target contributes 5 sketch variants as queries. Each query is paired with its corresponding high-fidelity diagram as the positive example, and with up to two randomly sampled diagrams from the candidate pool as negative examples.

\textbf{Implementation Details.} 
We implement our structure-aware retrieval framework using the Qwen3-VL-Embedding-2B~\cite{li2026qwen3} backbone. To efficiently align the sketch and diagram embedding spaces, we adopt Low-Rank Adaptation (LoRA). Specifically, the sketch and diagram encoders share a single set of LoRA weights, while the original parameters of the base model remain strictly frozen. During the fine-tuning process, the model is trained for 50 epochs using a peak learning rate of $1\times10^{-5}$ with a cosine annealing schedule and a 5\% linear warmup. To stabilize the optimization process, we apply gradient accumulation over 3 micro-batches before each optimizer update. All training experiments are conducted on a single node equipped with a NVIDIA H800 GPU with 80GB memory, utilizing CUDA version 12.4.

\textbf{Baselines.}
To comprehensively evaluate our framework, we divide our baselines into two distinct tracks: retrieval and generation. (1) \textbf{Retrieval Baselines.} To evaluate the effectiveness of our structure-aware retrieval approach, we compare our adapted model with several strong pre-trained vision models that are used without LoRA-based adaptation. The baseline models comprise OpenCLIP~\cite{openclip}, SIGLIP~\cite{zhai2023sigmoid}, SIGLIP2~\cite{tschannen2025siglip}, and the base Qwen3~\cite{li2026qwen3} model. We also introduce SKetch\_LVM~\cite{sain2023clip} as a sketch-specific embedding baseline, with its task-specific fine-tuning details provided in Appendix ~\ref{app:adde}. These models rely strictly on their original embedding spaces to retrieve candidates, highlighting the necessity of our topology-aware alignment. (2) \textbf{Generation Baselines.} To evaluate the final visual quality of the synthesized diagrams, we compare our retrieval-augmented generation pipeline against state-of-the-art generative models. The baselines include text-mediated generation frameworks, namely \texttt{AutoFigure} and \texttt{PaperBanana}. Furthermore, we compare against powerful end-to-end multimodal generators directly conditioned on sketches, including GPT-Image-2 and NanoBanana.
\textbf{Evaluation Metrics.} We evaluate our framework along three principal dimensions. (1) \textbf{Retrieval Metrics}: We report Mean Reciprocal Rank (MRR), Accuracy, Recall@1, Recall@5, and F1-score to quantitatively assess the degree of structural correspondence. (2) \textbf{Generation Metrics}: Conventional image quality metrics (e.g., FID) are often insufficient to characterize the stringent requirements for logical and semantic correctness in scientific diagrams. Consequently, we adopt a Vision-Language Model-as-a-Judge (VLM-as-a-Judge) evaluation protocol, which assesses eight fine-grained criteria encompassing content fidelity, communicative effectiveness, and visual design quality.(3) \textbf{Efficiency Metrics}: We report the end-to-end inference latency and the API cost on a per-sample basis. Formal definitions of these metrics, along with a detailed justification of our evaluation methodology, are provided in Appendix~\ref{app:de_me}.

\subsection{Results}

\textbf{Generation Quality.} Table~\ref{tab:gen} reports the VLM-as-a-Judge evaluation outcomes across the eight fine-grained sub-metrics as well as the aggregated overall score. The direct end-to-end generation models (GPT-Image-2 and Nano-Banana) exhibit limitations in \textit{Accuracy} and \textit{Completeness} criteria. In the absence of externally imposed structural priors, these models exhibit a tendency to hallucinate spurious connections or to omit critical methodological components that are represented in the sparse diagrammatic inputs. Text-mediated multi-agent baselines \texttt{AutoFigure} and \texttt{PaperBanana} demonstrate substantially superior performance, with \texttt{AutoFigure} attaining a strong overall score of 7.060. However, their dependence on intermediate textual or code-based representations frequently results in diminished fidelity in fine-grained layout control.

In contrast, our proposed retrieval-augmented frameworks, \texttt{DiagramRAG (NanoBanana)} and \texttt{DiagramRAG (GPT)}, attain the highest overall performance scores of 7.170 and 7.120, respectively. By conditioning the multimodal generators on retrieved, structurally compatible reference diagrams, \texttt{DiagramRAG} demonstrates particularly strong performance in \textit{Aesthetic Quality} (reaching up to 8.200) and \textit{Professional Polish} (8.200). The retrieved diagrams function as high-quality visual anchors, guiding the generator to synthesize refined, publication-ready illustrations while preserving the structural fidelity of the original user sketch.

\begin{table}[h]
\caption{VLM-as-a-Judge evaluation of diagram generation quality.}
\label{tab:gen}
\setlength{\heavyrulewidth}{1.5pt} 
\small
\setlength{\tabcolsep}{2.5pt}
\centering
\resizebox{\columnwidth}{!}{
\begin{tabular}{l|ccccccccc}
\toprule
\textbf{Method} & \multicolumn{1}{l}{\textbf{Aesthetic}} & \multicolumn{1}{l}{\textbf{Express}} & \multicolumn{1}{l}{\textbf{Polish}} & \multicolumn{1}{l}{\textbf{Clarity}} & \multicolumn{1}{l}{\textbf{Flow}} & \multicolumn{1}{l}{\textbf{Accuracy}} & \multicolumn{1}{l}{\textbf{Complete.}} & \multicolumn{1}{l}{\textbf{Approp.}} & \multicolumn{1}{l}{\textbf{overall}} \\ \hline
Paperbanana & 6.625 & 5.625 & 6.625 & 7.125 & 6.375 & 5.750 & 5.625 & 5.500 & 6.188 \\
Autofigure & 7.900 & 6.800 & 7.900 & \cellcolor[HTML]{EFEFEF}\underline{7.800} & 6.900 & 6.600 & 6.500 & 6.800 & 7.060 \\
GPT-image-2 & 7.600 & 6.400 & 7.600 & \cellcolor[HTML]{E4E4E4}\textbf{8.000} & \cellcolor[HTML]{EFEFEF}\underline{ 7.000} & \cellcolor[HTML]{EFEFEF}\underline{6.700} & 6.400 & 6.400 & 6.920 \\
NanoBanana & 6.500 & 5.500 & 6.500 & 7.400 & \cellcolor[HTML]{E4E4E4}\textbf{7.400} & 6.600 & 6.100 & 5.500 & 6.500 \\
\textbf{DiagramRAG(NanoBanana)} & \cellcolor[HTML]{EFEFEF}\underline{8.000} & \cellcolor[HTML]{EFEFEF}\underline{7.100} & \cellcolor[HTML]{EFEFEF}\underline{8.000} & 7.400 & 6.900 & \cellcolor[HTML]{E4E4E4}\textbf{6.704} & \cellcolor[HTML]{E4E4E4}\textbf{6.700} & \cellcolor[HTML]{E4E4E4}\textbf{7.101} & \cellcolor[HTML]{E4E4E4}\textbf{7.170} \\
\textbf{DiagramRAG(GPT)} & \cellcolor[HTML]{E4E4E4}\textbf{8.200} & \cellcolor[HTML]{E4E4E4}\textbf{7.105} & \cellcolor[HTML]{E4E4E4}\textbf{8.200} & 7.300 & 6.500 & 6.400 & \cellcolor[HTML]{EFEFEF}\underline{6.502} & \cellcolor[HTML]{EFEFEF}\underline{7.100} & \cellcolor[HTML]{EFEFEF}\underline{7.120} \\ 
\bottomrule
\end{tabular}
}
\end{table}

\textbf{Retrieval Performance.} As reported in Table~\ref{tab:rag}, traditional contrastive models (e.g., OpenCLIP), as well as the fully supervised, sketch-specific baseline Sketch-LVM (see Appendix~\ref{app:adde}), exhibit markedly limited performance, attaining F1-scores below 0.14 on DiagramBank. These results indicate a substantial domain shift and underscore their limited capacity to model the dense topological structures characteristic of this dataset. Conversely, our \textbf{Qwen3 + LoRA} configuration consistently outperforms all baseline models, attaining a Recall@5 of 0.894 on DiagramBank and an F1-score of 0.802 on FigureBench. These performance improvements substantiate our central hypothesis that effectively narrowing the representational gap between abstract sketches and publication-ready diagrams necessitates the combined use of a vision–language model’s intrinsic semantic reasoning capabilities and our explicit structure-aware enhancement mechanisms.
\begin{table}[h]
\setlength{\heavyrulewidth}{1.5pt} 
\caption{Quantitative comparison of structure-aware retrieval performance.}
\label{tab:rag}
\resizebox{\columnwidth}{!}{
\begin{tabular}{l|l|ccccc}
\toprule
\centering
\textbf{Dataset} & \textbf{Method} & \multicolumn{1}{l}{\textbf{MRR}} & \multicolumn{1}{l}{\textbf{Accuracy}} & \multicolumn{1}{l}{\textbf{Recall@1}} & \multicolumn{1}{l}{\textbf{Recall@5}} & \multicolumn{1}{l}{\textbf{F1-score}} \\ \hline
 & OpenCLIP & \multicolumn{1}{l}{0.113} & 0.077 & 0.077 & 0.132 & 0.077 \\
 & SKetch\_LVM & 0.150 & 0.107 & 0.107 & 0.178 & 0.107 \\
 & SIGLIP & \multicolumn{1}{l}{0.318} & 0.249 & 0.249 & 0.388 & 0.249 \\
 & SIGLIP2 & 0.453 & 0.386 & 0.386 & 0.513 & 0.386 \\
 & Qwen3 & \cellcolor[HTML]{EFEFEF}\underline{0.856} & \cellcolor[HTML]{EFEFEF}\underline{0.836} & \cellcolor[HTML]{EFEFEF}\underline{0.836} & \cellcolor[HTML]{EFEFEF}\underline{0.875} & \cellcolor[HTML]{EFEFEF}\underline{0.836} \\
\multirow{-6}{*}{DiagramBank} & \textbf{Qwen3 + LoRA} & \cellcolor[HTML]{E4E4E4}\textbf{0.870} & \cellcolor[HTML]{E4E4E4}\textbf{0.848} & \cellcolor[HTML]{E4E4E4}\textbf{0.848} & \cellcolor[HTML]{E4E4E4}\textbf{0.894} & \cellcolor[HTML]{E4E4E4}\textbf{0.848} \\ \hline
 & OpenCLIP & 0.124 & 0.095 & 0.095 & 0.137 & 0.095 \\
 & SKetch\_LVM & 0.161 & 0.130 & 0.130 & 0.193 & 0.130 \\
 & SIGLIP & 0.256 & 0.206 & 0.206 & 0.302 & 0.206 \\
 & SIGLIP2 & 0.333 & 0.282 & 0.282 & 0.378 & 0.282 \\
 & Qwen3 & \cellcolor[HTML]{EFEFEF}\underline{0.776} & \cellcolor[HTML]{EFEFEF}\underline{0.746} & \cellcolor[HTML]{EFEFEF}\underline{0.746} & \cellcolor[HTML]{EFEFEF}\underline{0.809} & \cellcolor[HTML]{EFEFEF}\underline{0.746} \\
\multirow{-6}{*}{FigureBench} & \textbf{Qwen3 + LoRA} & \cellcolor[HTML]{E4E4E4}\textbf{0.833} & \cellcolor[HTML]{E4E4E4}\textbf{0.802} & \cellcolor[HTML]{E4E4E4}\textbf{0.802} & \cellcolor[HTML]{E4E4E4}\textbf{0.874} & \cellcolor[HTML]{E4E4E4}\textbf{0.802} \\ 
\bottomrule
\end{tabular}
}
\end{table}

\textbf{Efficiency and Cost.} In addition to ensuring high-quality generation, a practical diagram illustration system must also exhibit computational efficiency. As detailed in Appendix~\ref{app:cost}, computationally intensive multi-agent frameworks such as \texttt{AutoFigure} require 12.62 minutes and \$0.400 per sample, primarily due to iterative coding cycles. In contrast, \texttt{DiagramRAG} substantially reduces online computation by pre-computing diagram embeddings into an offline library. \texttt{DiagramRAG (NanoBanana)} completes generation in 35.48 seconds at a cost of \$0.072 per sample. This corresponds to a 21$\times$ reduction in latency and an 82\% decrease in cost, demonstrating that the proposed framework exhibits high scalability and is well suited for deployment in real-world workflow scenarios.

\subsection{Ablation Study}

To better understand the source of performance gains in \texttt{DiagramRAG}, we conduct a comprehensive ablation study covering both the structure-aware supervision design in the retrieval stage and the agent-guided generation strategy in the generation stage.

\begin{figure}[!htb]
    \centering
    \captionsetup[subfigure]{skip=2pt}
    
    \begin{subfigure}{0.48\linewidth}
        \centering
        \includegraphics[
            width=\linewidth,
            trim={0.2cm 1.0cm 0.2cm 0.2cm},
            clip
        ]{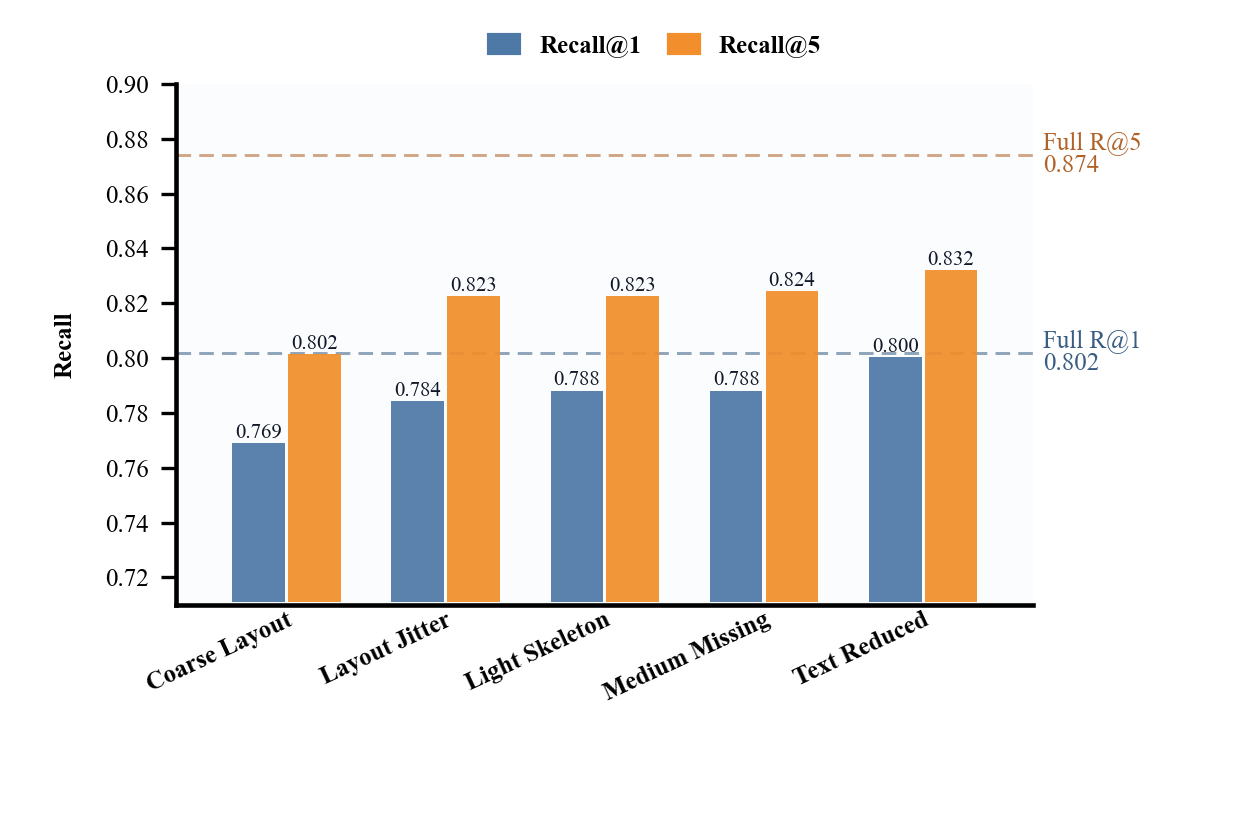}
        \caption{FigureBench}
        \label{fig:fgb_ab}
    \end{subfigure}
    \hfill
    \begin{subfigure}{0.48\linewidth}
        \centering
        \includegraphics[
            width=\linewidth,
            trim={0.2cm 1.0cm 0.2cm 0.2cm},
            clip
        ]{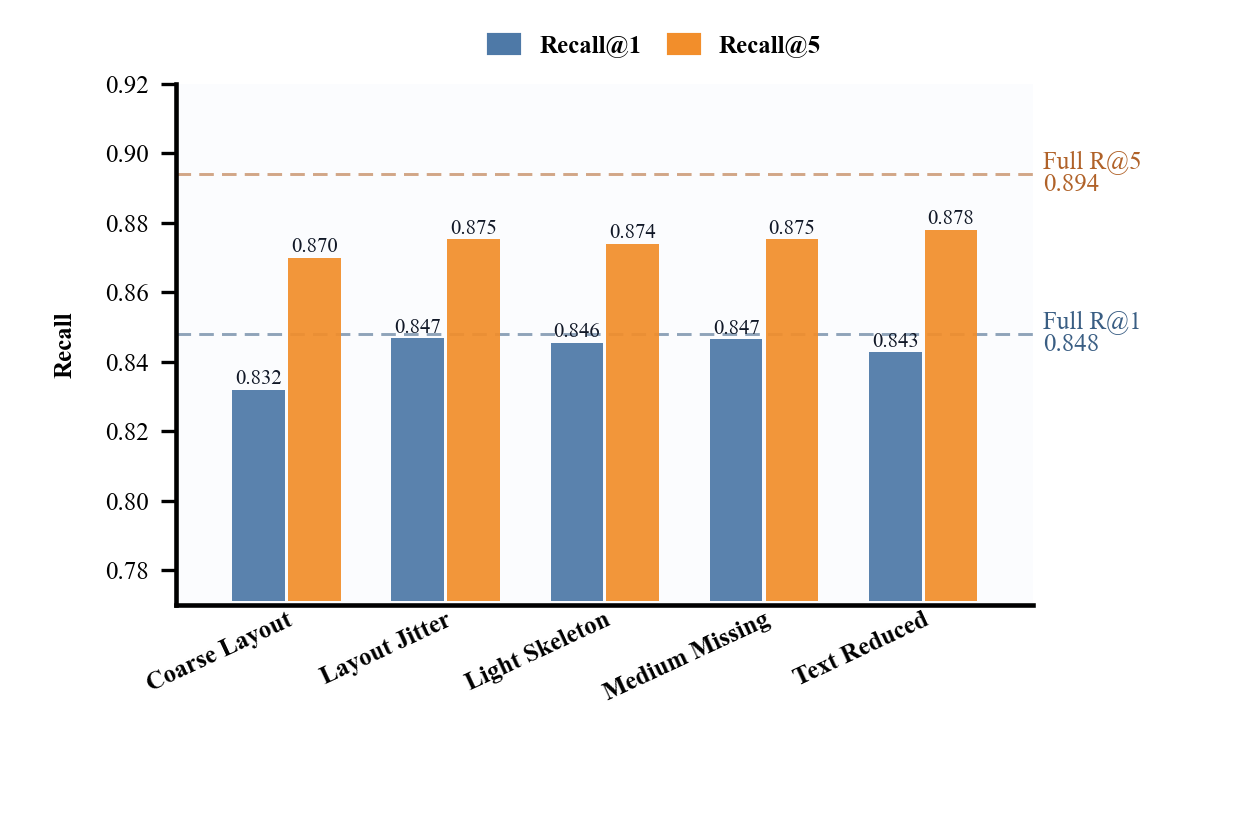}
        \caption{DiagramBank}
        \label{fig:dgb_ab}
    \end{subfigure}

    \vspace{-0.5em}
    \caption{Effect of sketch variants on retrieval performance.}
    \label{fig:ab_var}
    \vspace{-0.8em}
\end{figure}

\textbf{Effect of Sketch Variants.} To assess the contribution of each component within our heterogeneous topological supervision scheme, we perform a leave-one-variant-out ablation study. Beginning with the complete training configuration, we iteratively remove a single sketch-synthesis variant while holding the model architecture, retrieval corpus, and evaluation splits strictly fixed. As illustrated in Figure ~\ref{fig:ab_var}, the exclusion of any individual variant consistently results in a reduction of retrieval performance on both FigureBench and DiagramBank.

Specifically, removing the \textit{coarse layout} variant—which exhibits the most severe structural abstraction—leads to the most precipitous performance drop. On FigureBench, its removal causes the Recall@1 to plummet from 0.802 to 0.769, and the Recall@5 from 0.874 to 0.802. This substantial degradation demonstrates that exposing the model to heavily abstracted samples is crucial for forcing it to learn robust, topology-level correspondences. Furthermore, the absence of the \textit{text reduced} variant also noticeably harms performance, reducing the Recall@5 to 0.832 on FigureBench. This empirically confirms our hypothesis: without text-suppressed supervision, the embedding space risks collapsing into trivial semantic text-matching. By contrast, our full variant set compels the model to attend to layout, node relations, and connector geometries. Ultimately, the full model consistently achieves the upper bound across all metrics, proving that diverse structural degradation is essential for generalizing to the sparse and noisy nature of human-drawn sketches.


\paragraph{Impact of Dual-Agent Constraints.}
To evaluate the role of the two generation agents in the refined diagram generation stage, we conduct a $2\times2$ ablation study on the Structural Planning Agent and the Visual Guidance Agent. We keep the input sketch, retrieval index, retrieved references, top-$k$ setting ($k=3$), and image generation model unchanged, and only remove the corresponding agent. The full model achieves the highest overall score of 7.275. Removing the Structural Planning Agent decreases the overall score to 7.112, mainly due to drops in accuracy and completeness, indicating that structural planning helps preserve module coverage, connection relations, and topological consistency. In contrast, removing the Visual Guidance Agent leads to a larger degradation, reducing the overall score to 6.775, with particularly clear drops in aesthetic quality, visual expressiveness, and professional polish. Removing both agents further decreases the overall score to 6.750. These results suggest that the two prompt agents provide complementary constraints during generation: the Structural Planning Agent primarily improves structural fidelity, while the Visual Guidance Agent enhances visual quality and publication-level diagram style.Detailed ablation settings and metric-level analysis are provided in Appendix~\ref{app:dual_agent_ablation}.

\section{conclusion}

This paper presents \texttt{DiagramRAG}, a lightweight retrieval-augmented framework for sketch-based scientific diagram completion. It first converts high-quality diagrams into knowledge graph representations, where semantic content and topological structure are organized in the explicit structured space. Based on these representations, \texttt{DiagramRAG} synthesizes simplified sketch variants and adopts contrastive learning to align sketches with compatible diagrams, enabling efficient retrieval of references that are both semantically relevant and topologically compatible. Finally, the retrieved references are used as content, topology, and visual priors to guide diagram generation, allowing \texttt{DiagramRAG} to produce scientific diagrams that are semantically enriched, topologically complete, and visually polished. Empirical results on \texttt{DiagramBank} and \texttt{FigureBench} further verify the effectiveness of this design, showing that retrieval grounded in both semantics and topology can provide useful priors for high-quality diagram generation while keeping the overall framework lightweight and efficient.

\bibliography{main}

\appendix
\section{Additional Experimental Details}
\label{app:adde}
\textbf{Training Details for Sketch-Specific Baseline (Sketch-LVM).}
In Section 4.2, we noted that the sketch-specific baseline, Sketch-LVM, was evaluated under full supervision on our datasets. To ensure a rigorously fair comparison, we implement and train Sketch-LVM following its visual prompt tuning paradigm. We utilize a CLIP ViT image encoder as the visual backbone. Unless otherwise stated, we adopt CLIP ViT-B/32 and introduce three learnable visual prompt tokens for each input branch. Specifically, we maintain two separate sets of visual prompts: one for sketch queries and the other for original diagram images. The CLIP backbone is kept frozen, with the exception of LayerNorm parameters, which are allowed to be updated during training.

By default, the model is trained for 5 epochs with a mini-batch size of 64. We optimize the trainable parameters using Adam. The learning rate is set to $1 \times 10^{-4}$ for the visual prompt parameters and $1 \times 10^{-6}$ for the trainable LayerNorm parameters. The random seed is fixed to 42 for reproducibility. When using the larger CLIP ViT-L/14 backbone, we reduce the mini-batch size to 32 due to increased memory consumption and train the model for 10 epochs.

Training samples are constructed from the training split by pairing each sketch query with its corresponding original diagram according to their \texttt{source\_diagram\_id}. We optimize the model using an in-batch contrastive objective with a temperature of $0.05$. Within each mini-batch, diagrams associated with the same source diagram ID are considered positive samples, whereas diagrams from different source diagrams are treated as negatives. The objective is applied symmetrically in both sketch-to-diagram and diagram-to-sketch directions. 

To account for the varying difficulty of different sketch variants, we introduce variant-dependent loss weights. In the default setting, samples from the text-reduced variant (\texttt{text\_reduced\_force}) are assigned a weight of 4, samples from the coarse layout variant (\texttt{coarse\_layout}) are assigned a weight of 2, and all remaining variants are assigned a weight of 1. After each training epoch, we evaluate the model on the validation split using the same retrieval protocol as in testing. The model checkpoint with the highest validation MRR is selected for final evaluation, while the final-epoch checkpoint is retained only for reproducibility.

\section{Additional Results}
\label{app:addr}
\begin{table}[H]
\centering
\setlength{\heavyrulewidth}{1.5pt} 
\caption{End-to-end inference efficiency and computational cost.}
\label{tab:efficiency}
\begin{tabular}{lcc}
\toprule
Method & Time & Cost \\
\midrule
PaperBanana & 2.26 min & \$0.268 \\
AutoFigure & 12.62 min & \$0.400 \\
DiagramRAG (NanoBanana) & 35.48 s & \$0.072 \\
DiagramRAG (GPT) & 3.09 min & \$0.205 \\
\bottomrule
\end{tabular}
\end{table}
\subsection{End-to-End Inference Efficiency and Computational Cost.} 
\label{app:cost}
To further assess the practical deployability of \texttt{DiagramRAG}, we conduct a detailed quantitative evaluation of its inference latency and associated API costs for generating a single scientific illustration. As summarized in Table \ref{tab:efficiency}, multi-agent baseline methods that depend on iterative coding or multi-stage planning procedures (e.g., \texttt{AutoFigure} and \texttt{PaperBanana}) incur substantial temporal and financial overhead. In contrast, by precomputing diagram embeddings and storing them in an offline repository, \texttt{DiagramRAG} markedly reduces both latency and monetary expenditure, thereby enhancing the efficiency of the subsequent online structure-aware retrieval and generation pipeline.

\begin{figure}[h]
    \centering
    \includegraphics[width=0.65\linewidth]{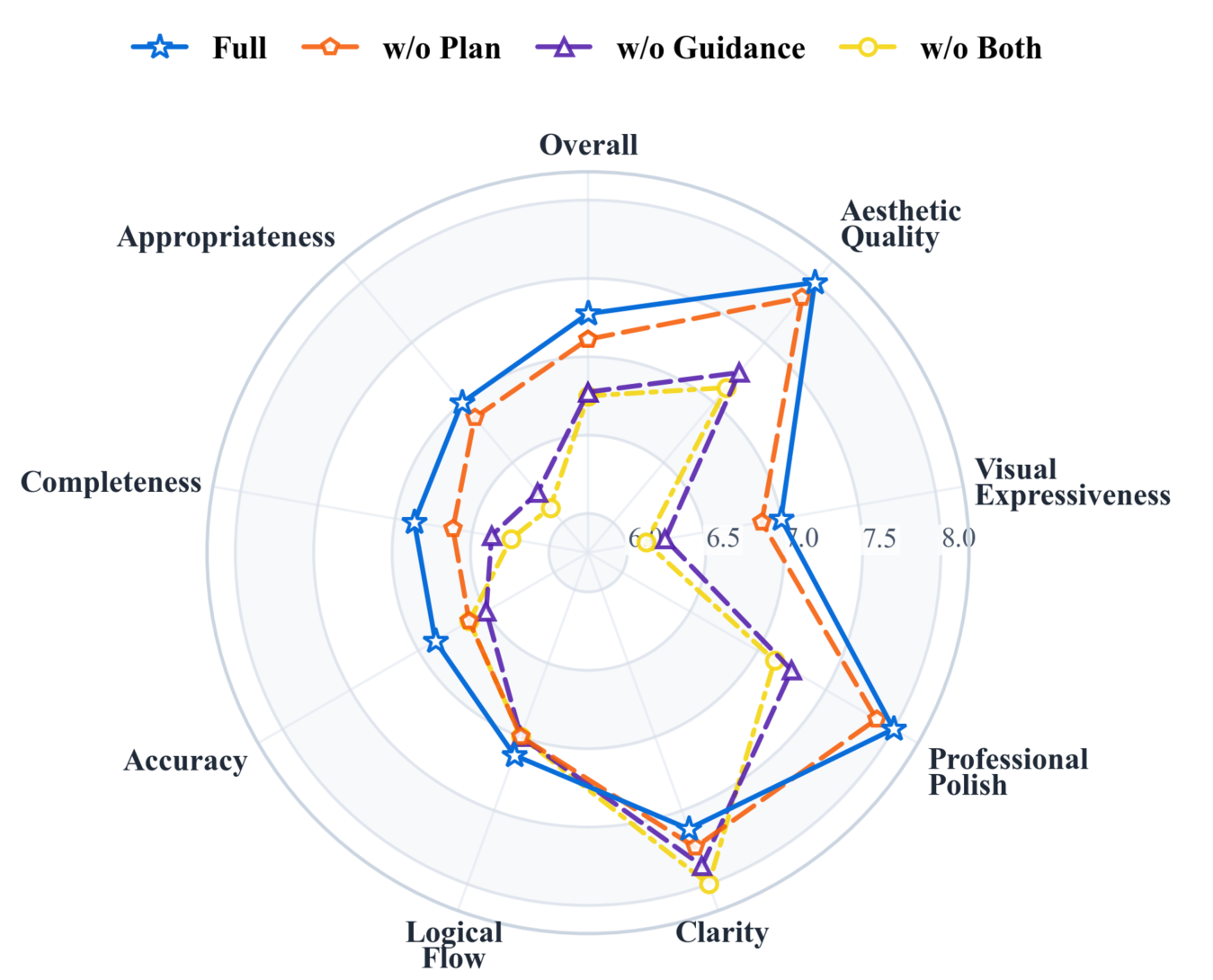}
    \caption{\textbf{Dual-Agent ablation.} VLM-based comparison of variants with and without the Structural Planning Agent and Visual Guidance Agent.}
    \label{fig:vlm_ablation_radar}
\end{figure}
\subsection{Dual-Agent Ablation Details}
\label{app:dual_agent_ablation}
To further analyze the contributions of structural planning and visual guidance in the refined diagram generation stage, we conduct a $2\times2$ agent ablation study. The full model enables both the Structural Planning Agent and the Visual Guidance Agent~\cite{shen2026brief}. The Structural Planning Agent treats the input sketch as the structural source of truth and explicitly constrains module coverage, grouping relations, arrow directions, and topological consistency. The Visual Guidance Agent uses the retrieved top-3 fine diagrams as visual references and guides layout organization, color logic, icon usage, local panels, feature/token/image thumbnails, and the level of detail expected in publication-quality scientific diagrams.

For a fair comparison, we keep the input sketch, retrieval index, retrieved references, top-$k$ setting ($k=3$), and image generation model unchanged across all variants, and only disable the corresponding agent instruction block. We compare four variants: the full model, w/o Structural Planning Agent, w/o Visual Guidance Agent, and w/o both agents. All variants are evaluated on the same samples using the same GPT-4o VLM judge.

As illustrated in Figure~\ref{fig:ab_var}, the full model achieves the highest overall score of 7.275. Removing the Structural Planning Agent decreases the overall score to 7.112, with both accuracy and completeness dropping by 0.25, indicating that structural planning helps preserve module coverage and topological consistency. In contrast, removing the Visual Guidance Agent leads to a larger degradation, reducing the overall score to 6.775. The drop is especially pronounced in aesthetic quality, visual expressiveness, and professional polish, each decreasing by 0.75. This suggests that visual guidance plays a key role in transferring publication-level visual cues, visual expressiveness, and detail density from the retrieved diagrams. When both agents are removed, the overall score further decreases to 6.750, with completeness and appropriateness dropping by 0.625 and 0.875, respectively.

Interestingly, the clarity score of the variant without both agents is slightly higher than that of the full model. We attribute this to the fact that, without structural planning and visual guidance, the generated diagrams tend to be simpler, contain fewer elements, and have lower visual density, which can make them appear clearer to the VLM judge. However, this apparent clarity does not translate into better overall quality: this variant obtains lower overall, completeness, appropriateness, and visual-design scores than the full model. These results show that the two agents provide complementary constraints during generation: the Structural Planning Agent mainly improves structural fidelity, while the Visual Guidance Agent primarily enhances visual quality and scientific-diagram style transfer.

\section{Details of Diagram Filtering}
\label{app:diagram_filtering}

The raw diagram collection extracted from scientific publications contains a large number of non-diagram images, including plots, tables, screenshots, natural images, and visually degraded samples. To construct the final high-quality diagram set $\mathcal{D}$, we apply a multi-stage filtering pipeline that combines low-level visual features, CLIP-based semantic alignment, supervised classification, and vision-language model ~\cite{zhang2026handling,zhang2025not}.

\paragraph{Low-level visual filtering.}
We first extract low-level visual features using conventional image processing techniques. Each image is converted to RGB format and resized for efficient feature computation. The extracted features describe visual quality and structural density, including resolution statistics, aspect ratio, colorfulness, saturation, color entropy, grayscale ratio, edge density, spatial frequency, image entropy, foreground ratio, and connected component count. These features capture whether an image contains sufficient visual content, edge structures, foreground regions, and layout complexity. We then apply coarse filtering rules to remove samples with clearly insufficient structural content or poor visual quality, such as images with very weak edge density or abnormal foreground proportion.

\paragraph{CLIP-based semantic quality representation.}
Since low-level visual features alone cannot determine whether an image is semantically suitable for \texttt{DiagramRAG}, we further use CLIP to estimate semantic alignment with target diagram characteristics. We construct positive textual concepts describing clear, well-organized, and readable scientific diagrams, and negative concepts describing confusing, crude, or illegible figures. For each image, we compute its cosine similarity to both positive and negative text prototypes and use their difference as a semantic quality margin:
\[
m_{\mathrm{clip}}
=
\frac{1}{N_+}\sum_i \cos(\mathbf{v}, \mathbf{t}^{+}_i)
-
\frac{1}{N_-}\sum_j \cos(\mathbf{v}, \mathbf{t}^{-}_j),
\]
where $\mathbf{v}$ denotes the image embedding, and $\mathbf{t}^{+}_i$ and $\mathbf{t}^{-}_j$ denote positive and negative text prototypes, respectively. A larger margin indicates stronger semantic alignment with high-quality structured diagrams.

\paragraph{Manual annotation and supervised retention modeling.}
To obtain task-specific supervision, we manually annotate a subset of candidate images as positive or negative examples. Positive samples are images that clearly match the target data distribution of \texttt{DiagramRAG}, such as method architecture diagrams, system diagrams, pipelines, workflows, flowcharts, or module interaction diagrams. Negative samples include figures that are not suitable for structure-aware modeling, such as experimental plots, tables, screenshots, natural images, qualitative result grids, dataset examples, pure text or formula figures, and images with poor visual quality or insufficient structural information. This annotation process provides a direct supervision signal for learning what should be retained in the final diagram set.

Based on the annotated data, we train a LightGBM binary classifier to jointly model visual quality and semantic relevance. For each annotated image, we construct a feature vector by concatenating low-level visual features and CLIP-derived semantic features:
\[
\mathbf{x} = [\mathbf{x}_{\mathrm{visual}}; \mathbf{x}_{\mathrm{clip}}].
\]
The low-level visual features describe basic image quality and structural density, including color statistics, edge density, spatial frequency, entropy, foreground ratio, and connected component count. The CLIP-derived features describe high-level semantic alignment, including similarities to positive and negative diagram descriptions and their semantic margin.

The classifier is trained to predict the manual keep/drop label:
\[
p_{\mathrm{keep}} = P(y=1 \mid \mathbf{x}),
\]
where $y=1$ denotes that the image should be retained. We use stratified cross-validation on the annotated subset to evaluate the classifier and select a reliable operating regime. After validation, the classifier is trained on the full annotated subset and applied to all candidate images to estimate their retention probabilities. In this way, the model learns a task-specific decision boundary from human judgments, while still leveraging both objective visual quality signals and semantic alignment signals.

\paragraph{Confidence-based stratification and VLM verification.}
After applying the trained LightGBM classifier to all candidate images, we stratify samples according to their predicted retention probabilities. Images with high $p_{\mathrm{keep}}$ are treated as high-confidence accepted samples, while those with low $p_{\mathrm{keep}}$ are treated as high-confidence rejected samples. Images whose probabilities fall between the two thresholds are considered uncertain, since they are close to the learned decision boundary and are more likely to contain ambiguous cases.

For these uncertain samples, we further employ a vision-language model to perform fine-grained verification under explicit task criteria~\cite{zhang2025harnessing}. The VLM is instructed to retain structured diagram types suitable for \texttt{DiagramRAG}, including method architecture diagrams, system diagrams, pipelines, workflows, flowcharts, and module interaction diagrams, and to reject plots, tables, screenshots, natural images, qualitative result grids, dataset examples, pure text/formula figures, and low-quality images.

\paragraph{Final decision fusion.}
The final decision is obtained by combining the high-confidence LightGBM decisions with the VLM verification results for uncertain samples. If the VLM fails to provide a reliable decision, the sample is rejected by default. This conservative strategy prioritizes precision over recall, ensuring that the resulting diagram set $\mathcal{D}$ contains images with clear structure, consistent semantic relevance, and strong suitability for downstream \texttt{DiagramRAG} tasks.

\section{Details of Topology-KG Variant Generation Protocol}
\label{app:kg_variant_generation}

We generate controlled topology-KG variants from diagram KG records. Each record contains a diagram identifier, an image path, metadata, and a graph field. The graph follows a topology-first schema with five components: nodes, directed edges, groups, and layout attributes. A node stores an identifier, name, semantic type, shape, normalized bounding box, center location, group assignment, and an importance tag. An edge stores source and target node identifiers, a path type, and a direction. Groups store member node identifiers and optional bounding boxes. The layout field records global topology attributes, including flow direction, topology type, reading order, and a short description of the main structure.

Before variant generation, each KG is normalized and cleaned. Invalid nodes without identifiers are removed. Missing node labels are normalized, bounding boxes and node centers are clipped to the unit image coordinate range, containment edges are assigned type \texttt{containment}, and duplicate edges are removed. Group memberships are filtered to valid node identifiers. Layout lists such as main path, branch points, merge points, reading order, and feedback edges are restricted to retained nodes. If branch points, merge points, reading order, or main path are missing after filtering, they are inferred from node positions and non-containment edges. Group bounding boxes are recomputed from their retained member nodes when possible.

For each KG \(G_i\), we construct five deterministic variant instances:
\textsc{LightSkeleton}, \textsc{MediumMissing}, \textsc{CoarseLayout},
\textsc{TextReduced}, and \textsc{LayoutJitter}.
The random seed associated with each pair \((G_i, v)\) is obtained by hashing the global seed, the identifier of the source diagram, and the variant name. This procedure guarantees that all stochastic operations, including layout jitter, are fully reproducible.

\paragraph{Text reduction policy.}
All variants apply constraints on textual content to ensure that the degraded KG cannot preserve the full original labels. Three levels of text retention are defined. At the \texttt{medium} level, node labels are preserved but truncated to at most three words; group labels are preserved up to a maximum of two words; and at most a single textual element is retained from among role title, label, legend, or annotation, truncated to two words. At the \texttt{low} level, only short node and group labels are retained, and all free-text elements are removed. At the \texttt{none} level, node labels, group labels, layout text, and all free-text elements are removed, with the exception that input and output nodes may be replaced by generic placeholder labels.

\paragraph{\textsc{LightSkeleton}.}
This variant removes only nodes whose importance is marked as \texttt{decorative}. Edges are retained only when both endpoints survive; no bridge edges are added through removed nodes. Groups and layout attributes are cleaned according to the retained node set. The \texttt{medium} text policy is then applied. Thus, \textsc{LightSkeleton} preserves the main topology while deleting visually decorative elements and shortening text.

\paragraph{\textsc{MediumMissing}.}
This variant keeps structural anchors rather than all non-decorative nodes. The retained set includes nodes appearing in the main path, branch points, merge points, or feedback edges, as well as nodes whose type is input, output, container, connector, branch, or merge. If this set is empty, major-importance nodes are used as a fallback. Removed nodes may be bridged: when a path from one retained node to another passes only through removed nodes, a new edge is inserted and marked with its original intermediate path using a \texttt{bridged\_from} field. Free text items are removed before the \texttt{medium} text policy is applied. This variant simulates missing intermediate content while preserving global connectivity.

\paragraph{\textsc{CoarseLayout}.}
This is the strongest structural simplification. It keeps topology-critical anchors, including branch points, merge points, feedback endpoints, input nodes, output nodes, and a sparse sample of the main path. The main path sample preserves endpoints and targets approximately $22\%$ of the path, with at most six sampled nodes. Only structurally useful connectors with degree at least three are kept. If no explicit topology anchors are available, the method falls back to a small layout-ordered sample of major nodes. A node budget is then enforced: the retained set is limited to at most $30\%$ of the original nodes and at most six nodes, while preserving at least three nodes when possible and restoring main-path endpoints if needed. Text is reduced with the \texttt{low} policy. This variant keeps a coarse visual backbone rather than the full detailed topology.

\paragraph{\textsc{TextReduced}.}
This variant first applies \textsc{LightSkeleton}, then removes nearly all semantic text. Input and output nodes are renamed to generic labels \texttt{Input} and \texttt{Output}; all other node names, group labels, free text items, and the layout main-structure description are removed. The graph topology and retained bounding boxes are otherwise preserved from the light skeleton.

\paragraph{\textsc{LayoutJitter}.}
This variant first applies \textsc{MediumMissing}, then perturbs the bounding box of each retained node. The box center is shifted independently in $x$ and $y$ by a uniform offset in $[-\rho,\rho]$, where $\rho$ is the jitter fraction and defaults to $0.05$. Width and height are scaled by independent factors sampled from $[1-0.8\rho,1+0.8\rho]$. The resulting boxes are clipped to the unit coordinate range, node centers are recomputed, and group bounding boxes are refreshed with padding. The \texttt{medium} text policy is applied after jittering.

\begin{algorithm}[H]
\caption{Canonical Topology-KG $\rightarrow$ Variant KG Generation}
\label{alg:kg_variants}
\begin{algorithmic}[1]
\raggedright
\Require canonical KG records $\mathcal{R}$; variant set $\mathcal{V}$; global seed $s_0$; jitter fraction $\rho$
\Ensure variant KG records $\{R_i^v\}$ and variant loss records $\{\mathcal{L}_i^v\}$
\For{each record $R_i \in \mathcal{R}$}
  \State Obtain source identifier $d_i$ from the image path or diagram identifier;
  \State Extract graph $G_i$ and normalize nodes, edges, groups, texts, and layout fields;
  \State Remove invalid nodes, invalid edges, self-loops, and duplicate edges;
  \State Filter group memberships and layout references to valid retained nodes;
  \State Infer missing main path, reading order, branch points, merge points, and flow direction when needed;
  \For{each variant $v \in \mathcal{V}$}
    \State Initialize deterministic seed $r_{i,v}\gets\textsc{StableSeed}(s_0,d_i,v)$;
    \If{$v=\textsc{LightSkeleton}$}
      \State Remove decorative nodes and retain edges only between surviving nodes;
      \State Apply the medium text policy;
    \ElsIf{$v=\textsc{MediumMissing}$}
      \State Keep structural layout anchors and typed structural nodes;
      \State Add bridge edges through removed intermediate nodes when possible;
      \State Apply the medium text policy;
    \ElsIf{$v=\textsc{CoarseLayout}$}
      \State Keep branch/merge points, feedback endpoints, endpoints, useful connectors, and a sparse main-path sample;
      \State Enforce the coarse node budget and apply the low text policy;
    \ElsIf{$v=\textsc{TextReduced}$}
      \State Start from \textsc{LightSkeleton};
      \State Replace input/output labels with generic labels and remove other text;
    \ElsIf{$v=\textsc{LayoutJitter}$}
      \State Start from \textsc{MediumMissing};
      \State Perturb retained node bounding boxes using seed $r_{i,v}$ and jitter fraction $\rho$;
      \State Refresh group bounding boxes and apply the medium text policy;
    \EndIf
    \State Clean the resulting graph $G_i^v$;
    \State Compute loss record $\mathcal{L}_i^v\gets\textsc{VariantLoss}(G_i,G_i^v)$;
    \State Store a variant record with source identifier $d_i$, variant name $v$, graph $G_i^v$, and $\mathcal{L}_i^v$;
  \EndFor
\EndFor
\end{algorithmic}
\end{algorithm}
\section{Degradation Statistics of KG Variants}
\label{app:kg_variant_losses}
To quantitatively evaluate the extent of information removed or modified during the generation of knowledge graph (KG) variants, we compute a set of degradation metrics for each original graph \(G_i\) and its corresponding variant \(G_i^v\). These metrics are designed to characterize three primary forms of degradation: node loss, edge loss, and text loss.

\begin{figure}[H]
  \centering
  \includegraphics[width=\linewidth]{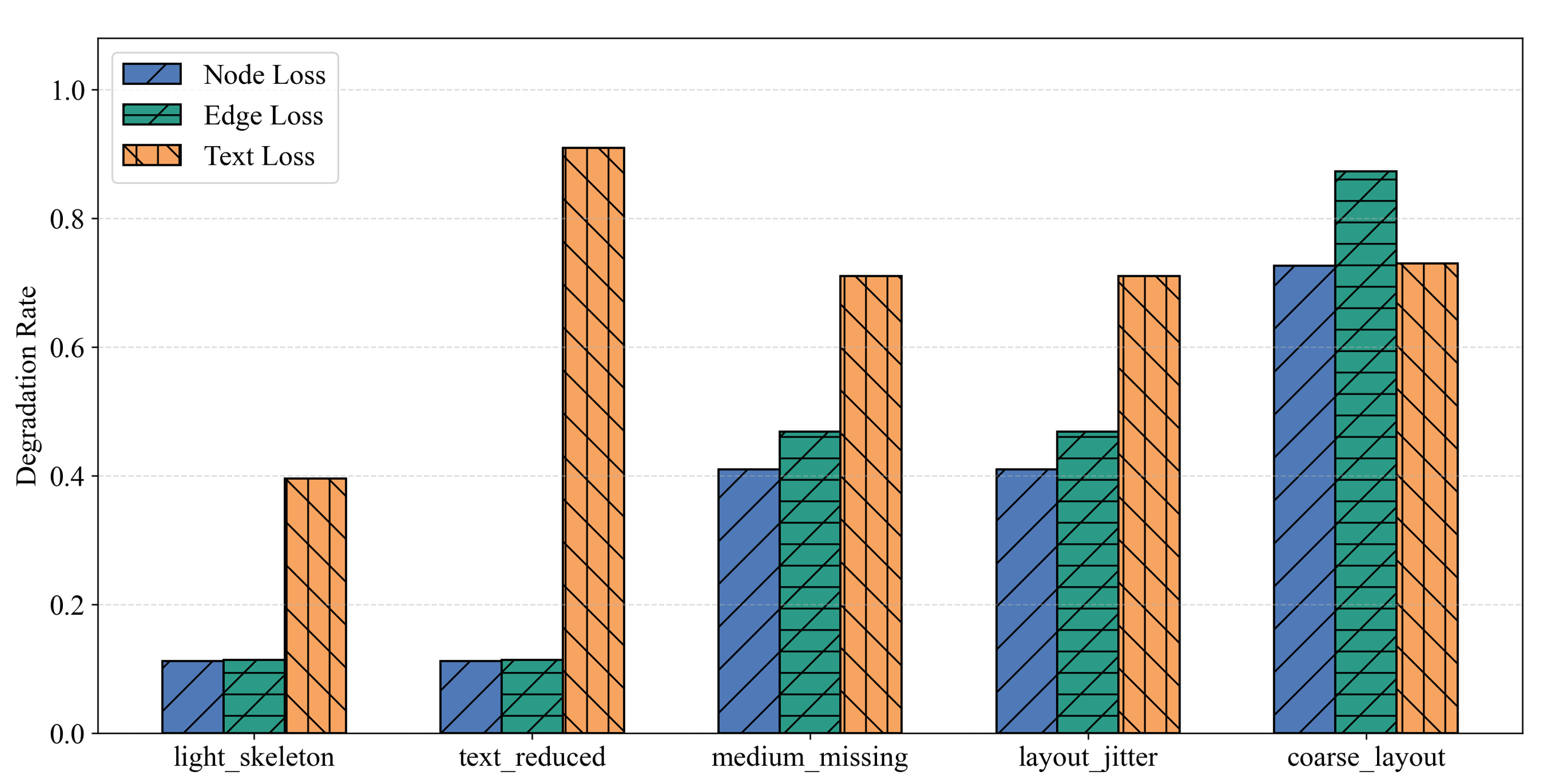}
  \caption{Mean information loss in the Diagrambank dataset.}
  \label{fig:dgbloss}
\end{figure}

\begin{figure}[H]
  \centering
  \includegraphics[width=\linewidth]{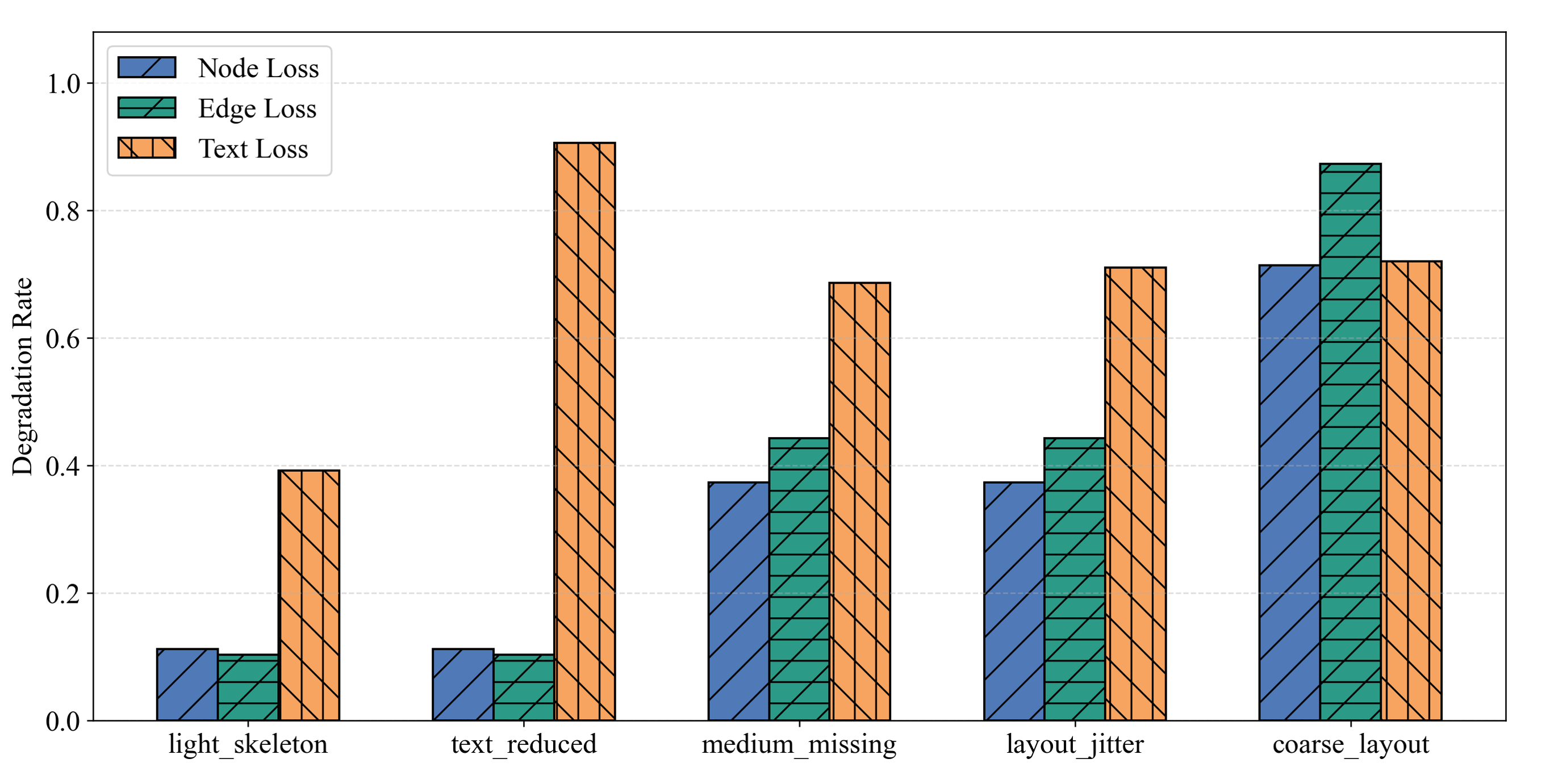}
  \caption{Mean information loss in the Figurebench dataset.}
  \label{fig:fgbloss}
\end{figure}

\section{Detailed Evaluation Metrics}
\label{app:de_me}
To complement Section 4.1, we provide a detailed description of our evaluation metrics and the rationale behind our generation assessment protocol.

\textbf{Retrieval Metrics.} To quantitatively assess the capability of the models to retrieve structurally compatible diagrams from the candidate pool, we report Mean Reciprocal Rank (MRR), Accuracy, Recall@1, Recall@5, and F1-score. Collectively, these metrics provide a rigorous evaluation of the correspondence between the sparse input sketch and the dense diagram topology.

\textbf{Generation Metrics.} In accordance with evaluation paradigms recently introduced by automated scientific illustration systems such as \texttt{AutoFigure} and \texttt{PaperBanana}, we omit conventional image-level metrics (e.g., FID), as these measures are frequently misaligned with the stringent logical and topological correctness requirements of scientific diagrams. Instead, we adopt a fully VLM-as-a-Judge–based evaluation protocol. In this framework, a vision-language model (VLM) quantitatively assesses the generated diagrams with respect to human-authored, ground-truth figures along three primary dimensions:
\begin{itemize}
    \item \textit{Content Fidelity}: Measures adherence to the underlying source context, operationalized in terms of accuracy, completeness, and contextual appropriateness.
    \item \textit{Communication Effectiveness}: Evaluates the degree to which the diagram facilitates comprehension, characterized by clarity and logical flow.
    \item \textit{Visual Design}: Captures aesthetic and professional qualities, assessed through aesthetic quality, visual expressiveness, and the level of professional polish.
\end{itemize}

\textbf{Efficiency Metrics.} To quantitatively characterize the trade-off between model performance and practical deployment feasibility, we additionally report the end-to-end inference time and the monetary cost per processed sample.

\section{Details of prompts}
\label{app:prompt}
\subsection{Prompts Overview}
\begin{appendixpromptbox}
The DiagramRAG pipeline employs six distinct prompts, each corresponding to a key processing stage: (1) vision–language model (VLM)-as-judge filtering, (2) knowledge graph extraction, (3) sketch generation, (4) structural planning, (5) visual guidance, and (6) VLM-based absolute evaluation. The full prompt templates are provided in \texttt{prompts.txt}.
\end{appendixpromptbox}
\subsection{VLM-as-judge filtering}
\begin{appendixpromptbox}
\paragraph{System Prompt.} You are evaluating whether a figure should be included in a high-quality scientific-diagram dataset for retrieval-augmented generation. The figure has already been pre-filtered and lies in the AMBIGUOUS middle of a probe model — your job is to make the final keep/drop call.

\paragraph{User Prompt.} 
\begin{Verbatim}[
  breaklines=true,
  breakanywhere=true,
  breaksymbolleft={},
  breaksymbolright={},
  fontsize=\small
]
KEEP if the figure is one of:
  - architecture / model / system diagram of the paper's proposed method
  - training pipeline or inference procedure of the proposed approach
  - workflow or flowchart describing the proposed method logic
  - module / block diagram showing concrete component interactions
DROP if the figure is:
  - experiment plot or chart (line / bar / scatter / radar / heatmap)
  - table
  - screenshot or UI capture
  - natural image / photo
  - qualitative result gallery or comparison grid
  - dataset / benchmark example
  - pure formula / pure text / legend-only figure
  - generic conceptual illustration without concrete method structure
  - blurry / low-resolution / broken / cropped figure
Decision rules:
  - Trust the image; ignore any caption.
  - For multi-panel figures, judge by the dominant panel.
  - If uncertain, err on DROP (precision > recall).
  - If decision is 'drop', label MUST be 'other'.
  - If decision is 'keep', label MUST be one of architecture | pipeline | workflow | flowchart | overview.
Return exactly ONE JSON object, no extra text:
{"decision": "keep|drop", "label": "...", "confidence": 0.00, "reason": "<one short phrase>"}
\end{Verbatim}
\end{appendixpromptbox}

\subsection{knowledge graph extracting}
\begin{appendixpromptbox}
\paragraph{System Prompt.} You are a meticulous topology extraction engine for scientific figures. Extract redrawable graph structure from the image only. Prioritize topological fidelity over semantic summarization. Return exactly one valid JSON object and no prose.

\paragraph{User Prompt.} 
Extract a redrawable topology graph from the research figure. The output should preserve enough structure for another program to reconstruct a sketch with the same topology, including repeated modules, branches, merges, containers or groups, main paths, reading order, and feedback loops.
Return a single JSON object with the following fields:
\begin{Verbatim}[
  breaklines=true,
  breakanywhere=true,
  breaksymbolleft={},
  breaksymbolright={},
  fontsize=\small
]
{
  "nodes": [
    {
      "id": "n1",
      "name": "visible label or neutral role",
      "type": "input|output|module|component|container|connector|branch|merge|stage|visual|text|other",
      "shape": "rectangle|circle|diamond|image_panel|table|text|container|point|line|other",
      "bbox": [x1, y1, x2, y2],
      "x": 0.5,
      "y": 0.5,
      "group": "optional group id",
      "importance": "major|minor|connector|decorative"
    }
  ],
  "edges": [
    {
      "source": "n1",
      "target": "n2"
    }
  ],
  "layout": {
    "flow_direction": "horizontal|vertical|mixed|unknown",
    "main_structure": "one sentence describing the visible topology"
  }
}
\end{Verbatim}
Preserve repeated visual modules as separate nodes, even when labels are identical. Add explicit connector, branch, or merge nodes when they affect topology. Preserve containers, edge directions, edge geometry, feedback loops, and the dominant information-flow path. Use normalized bounding boxes in $[0,1]$ with the origin at the top-left image corner. Use stable node IDs and reference only these IDs in edges, groups, paths, and layout fields. Use only information visible in the figure; do not invent scientific modules.
The output must be a single syntactically valid JSON object without markdown, code fences, explanations, or trailing commas.

\end{appendixpromptbox}

\subsection{sketch generating}
\begin{appendixpromptbox}
\paragraph{System Prompt.} You are a topology-preserving sketch generation assistant. Given a source topology render and an authoritative knowledge graph, redraw the diagram as a clean black-and-white hand-drawn line sketch. Preserve only the KG-supported nodes, labels, containers, and directed edges. Use the source image only as a weak layout reference. Do not invent or omit structural elements, do not copy visual style or exact geometry, and do not add icons, textures, colors, captions, or decorative marks.

\paragraph{User Prompt.} 
Draw a new topology line diagram on a pure white canvas. Use the attached source render only as a weak layout reference for coarse placement, ordering, and grouping. The KG is the authoritative drawing specification for all nodes, labels, containers, and directed edges.
Redraw the diagram as a clean black-and-white hand-drawn sketch using only simple geometric primitives such as boxes, circles, container outlines, arrows, and short handwritten labels. Do not copy the source image's exact geometry, text style, colors, icons, pictorial content, frame boundaries, or connector routes.
Preserve every major KG node and KG edge. Do not invent new nodes, labels, arrows, groups, or relationships. For each KG edge, the arrow tail must attach to the source node boundary and the arrowhead must attach to the target node boundary. Reroute arrows when needed to avoid overlaps while keeping the KG connectivity unchanged.
Keep the sketch sparse, readable, and structurally faithful. Use black pen-like strokes with slight hand-drawn irregularity. Avoid vector-like diagrams, printed typography, screenshots, decorative marks, textures, shading, captions, legends, and any non-KG visual elements.
Authoritative KG drawing plan:
\begin{Verbatim}[
  breaklines=true,
  breakanywhere=true,
  breaksymbolleft={},
  breaksymbolright={},
  fontsize=\small
]
<Minimal KG JSON. Use only these nodes and edges; do not infer missing visual styles:
{...}>
\end{Verbatim}
\end{appendixpromptbox}

\subsection{structural planning}
\begin{appendixpromptbox}
\paragraph{System Prompt.}You are a scientific diagram generation assistant. Treat the first image as the structural blueprint. Infer the main modules, grouping, data flow, and relationships from it, but do not copy its rough sketch style, node/edge IDs, debug labels, or placeholder text. Preserve the intended structure, not the visual appearance.
\paragraph{User Prompt.}Use the first image mainly as the semantic and structural plan: infer the main components, their grouping, data-flow direction, and high-level relationships. Do not imitate its rough drawing style, black-and-white simplicity, node IDs, edge IDs, or graph/debug annotations. If the sketch has placeholder labels, replace them with clean, concise academic module names.

\end{appendixpromptbox}

\subsection{visual guidance}
\begin{appendixpromptbox}
\paragraph{System Prompt.}You are a scientific diagram generation assistant.Treat the retrieved reference diagrams as visual style guidance. Use them for polished layout, spacing, colors, typography, icons, arrows, and paper-ready visual quality, but do not copy their exact content, paper-specific text, logos, or module names.
\paragraph{User Prompt.}Visual Guidance images follow. These are clean fine diagrams. Use them for layout polish, colors, icons, image panels, typography, spacing, arrows, and paper-ready visual style.
\end{appendixpromptbox}

\subsection{VLM-based absolute evaluation}
\begin{appendixpromptbox}
\paragraph{System Prompt.}
You are an impartial VLM-as-a-judge evaluator for scientific framework figures. Evaluate the candidate figure objectively. Use the sketch as the structural source of truth and the human-created reference figure as the target-quality reference. Return exactly one valid JSON object and no prose.

\paragraph{User Prompt.}
You are given three images:
Image 1 is the sketch-derived structural source of truth.
Image 2 is the human-created fine/reference figure for the same source.
Image 3 is the candidate generated figure to evaluate.

Evaluate Image 3 against Image 1 and Image 2. Score each sub-metric from 0 to 10. Use the sketch to judge topology, containment, arrow directions, and module coverage. Use the reference figure to judge publication quality and intended target content.

The sub-metrics are: aesthetic quality, visual expressiveness, professional polish, clarity, logical flow, accuracy, completeness, and appropriateness. Penalize unreadable text, copied reference-specific content from unrelated examples, invented titles or captions, missing or reversed arrows, dropped modules, irrelevant objects, and debug labels such as node IDs or edge IDs.

Return a single JSON object with the following fields:
\begin{Verbatim}[
  breaklines=true,
  breakanywhere=true,
  breaksymbolleft={},
  breaksymbolright={},
  fontsize=\small
]
{
  "scores": {
    "aesthetic_quality": 0,
    "visual_expressiveness": 0,
    "professional_polish": 0,
    "clarity": 0,
    "logical_flow": 0,
    "accuracy": 0,
    "completeness": 0,
    "appropriateness": 0
  },
  "overall": 0,
  "strengths": [],
  "weaknesses": [],
  "most_important_fix": ""
}
\end{Verbatim}
The output must be a single syntactically valid JSON object without markdown, code fences, explanations, or trailing commas.

\end{appendixpromptbox}

\section{Limitations} Although \texttt{DiagramRAG} improves sketch-based scientific diagram generation through structure-aware retrieval, it has several limitations. First, its retrieval quality depends on the coverage and quality of the curated diagram corpus; sketches whose topology or domain is poorly represented in the corpus may retrieve less suitable references. Second, the construction of sketch-diagram supervision relies on knowledge-graph extraction and sketch synthesis, which may introduce noise for diagrams with dense layouts, ambiguous connectors, or very small text. Third, the final generation quality still depends on external multimodal generators, which may occasionally produce visually plausible but structurally incorrect diagrams. Finally, our evaluation focuses on scientific framework and workflow diagrams, and the generalization of \texttt{DiagramRAG} to other figure types such as plots, tables, screenshots, or mathematical illustrations remains future work.


% ...

\newpage

\end{document}